\documentclass[conference]{IEEEtran}
\IEEEoverridecommandlockouts
\usepackage{cite}
\usepackage{placeins}
\usepackage{stfloats}
\usepackage{cuted}
\usepackage{caption}
\usepackage{amsmath,amssymb,amsfonts}

\usepackage{algorithmic}
\usepackage{algorithm}
\usepackage{graphicx}
\usepackage{textcomp}
\usepackage{xcolor}
\usepackage{booktabs}
\usepackage{subcaption}
\usepackage{multirow}
\usepackage{multicol}
\usepackage{array}
\usepackage[inline]{enumitem}

\def\BibTeX{{\rm B\kern-.05em{\sc i\kern-.025em b}\kern-.08em
    T\kern-.1667em\lower.7ex\hbox{E}\kern-.125emX}}

\begin{document}

\title{Words into World: A Task-Adaptive Agent for Language-Guided Spatial Retrieval in AR}

\author{\IEEEauthorblockN{Lixing Guo}
\IEEEauthorblockA{\textit{University of California, Santa Barbara}\\
Santa Barbara, California, USA\\
lixing\_guo@ucsb.edu}
\and
\IEEEauthorblockN{Tobias Höllerer}
\IEEEauthorblockA{\textit{University of California, Santa Barbara}\\
Santa Barbara, California, USA\\
holl@cs.ucsb.edu}
}

\maketitle

\begin{strip}
\centering
\vspace{0.5em}
\includegraphics[width=0.65\textwidth]{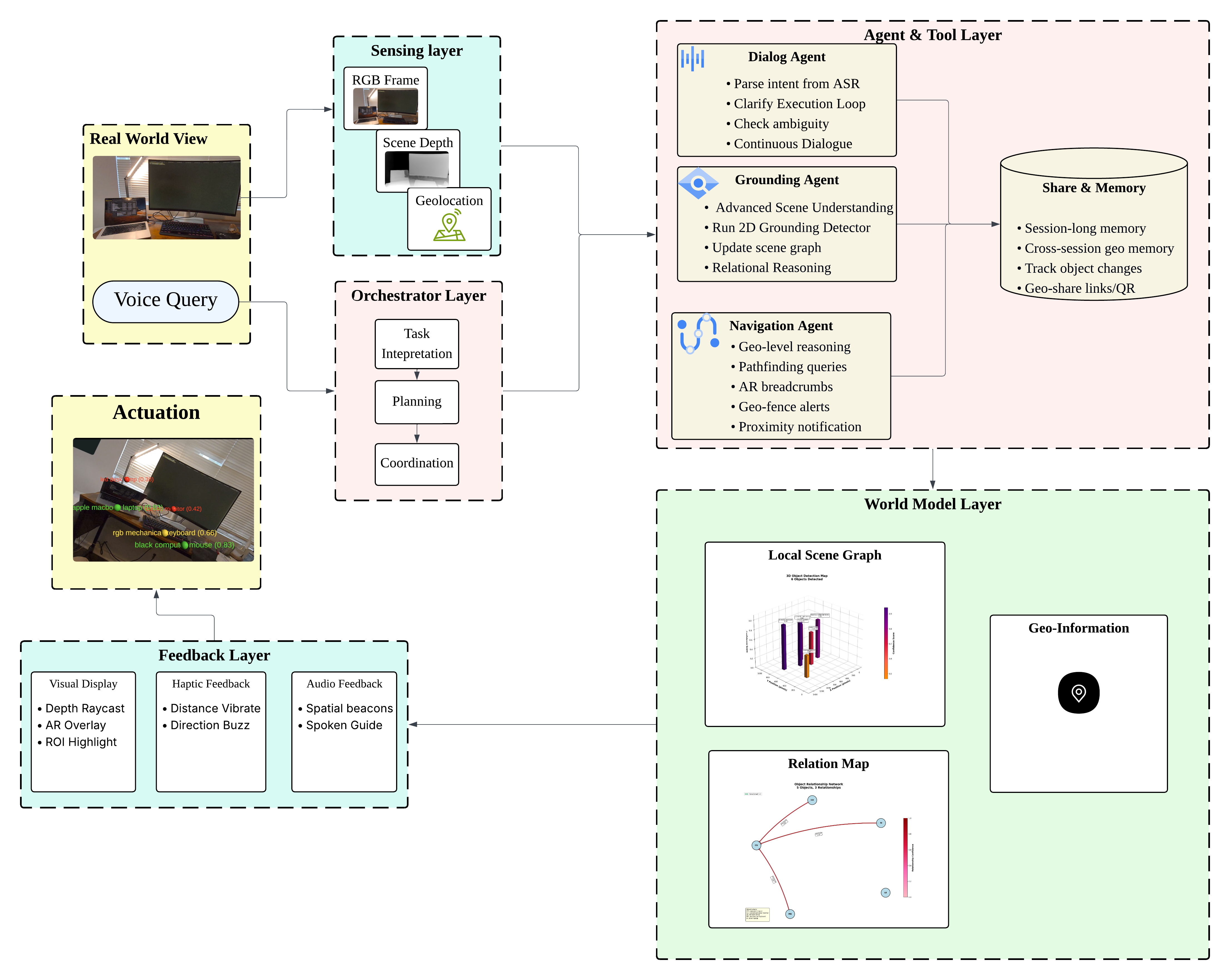}
\captionof{figure}{Overview of the \emph{Words into World} architecture. Voice queries and RGB views from Meta Quest~3 flow through five layers: \textbf{Sensing} captures passthrough RGB, depth, and geolocation; \textbf{Orchestrator} interprets tasks and coordinates agents; \textbf{Agent \& Tool} handles dialog, grounding, and navigation; \textbf{World Model} maintains scene graphs, relations, and geo-information; \textbf{Feedback} renders visual overlays, haptic cues, and audio guidance.}
\label{fig:system-architecture}
\vspace{0.5em}
\end{strip}

\begin{abstract}
Traditional augmented reality (AR) systems predominantly rely on fixed-class detectors or fiducial markers, limiting their ability to interpret complex, open-vocabulary natural language queries. We present a modular AR agent system that integrates multimodal large language models (MLLMs) with grounded vision models to enable relational reasoning in space and language-conditioned spatial retrieval in physical environments. Our adaptive task agent coordinates MLLMs and coordinate-aware perception tools to address varying query complexities, ranging from simple object identification to multi-object relational reasoning, while returning meter-accurate 3D anchors. It constructs dynamic AR scene graphs encoding nine typed relations (spatial, structural-semantic, causal-functional), enabling MLLMs to understand not just what objects exist, but how they relate and interact in 3D space. Through task-adaptive region-of-interest highlighting and contextual spatial retrieval, the system guides human attention to information-dense areas while supporting human-in-the-loop refinement. The agent dynamically invokes coordinate-aware tools for complex queries—selection, measurement, comparison, and actuation—grounding language understanding in physical operations. The modular architecture supports plug-and-use vision-language models without retraining, establishing AR agents as intermediaries that augment MLLMs with real-world spatial intelligence for interactive scene understanding. We also introduce GroundedAR-Bench, an evaluation framework for language-driven real world localization and relation grounding across diverse environments.
\end{abstract}


\section{Introduction}

People increasingly expect AR headsets to answer spatial questions about their surroundings, not just pin floating windows in mid-air. A user wearing a Meta Quest~3 or HoloLens~2 might look at a cluttered desk and ask, \emph{``What is on the desk within reach?''} or \emph{``Which box is behind the toolbox but closer to me than the printer?''}, and expect an AR agent to respond with precise, grounded answers in situ. Delivering such experiences requires more than overlaying labels on camera images: it demands agents that connect open-vocabulary language to metrically accurate 3D structure, reason about object--object relations, and operate interactively in live, everyday environments.

This vision is technically within reach yet practically elusive. Commodity headsets provide high-resolution passthrough, depth sensing, and spatial meshes, while multimodal large language models (MLLMs) offer open-vocabulary recognition and conversational interaction. However, current MLLMs are fundamentally 2D-centric: they infer semantics from image pixels but lack robust geometric awareness, depth estimation, or explicit reasoning over object--object relations. Recent evaluations show that when applied directly to AR scenes, these models often misjudge depth, confuse object configurations, or hallucinate spatial relations, especially under occlusion or clutter. As a result, AR systems that rely solely on 2D semantics struggle to support core tasks such as spatial retrieval (e.g., \emph{``find the TV controller on the shelf''}), structured relational grounding (e.g., \emph{``What objects between me and the door could cause me to trip if I walk there in the dark?''}), aesthetic coordination (e.g., \emph{``Which objects in this room should I remove because they do not fit the room's color theme?''}), safety assessment (e.g., \emph{``Which objects in this room could be potentially dangerous for a two-year-old child?''}), or in-situ interpretation of complex workspaces.

A growing body of work seeks to bridge semantics and 3D structure. Spatial-augmentation frameworks such as SAMR~\cite{samr} attach 3D geometry to objects by combining segmentation, raycasting, and mesh fitting, and multimodal fusion systems such as \emph{Multimodal 3D Fusion and In-Situ Learning for Spatially Aware AI}~\cite{10765411} fuse RGB, depth, and semantic embeddings into global volumetric scene representations for natural-language search and in-situ learning. Open-vocabulary object placement pipelines such as OCTO+~\cite{octoPlus} use RAM++, Grounding DINO, and GPT-4 to select semantically appropriate 2D placement locations for virtual content before raycasting into 3D. In parallel, in-situ learning systems like KELVAR~\cite{huynhKelvar} and AR authoring agents such as agentAR~\cite{agentAR} demonstrate that AR devices can support persistent object labels and LLM-driven tool orchestration in everyday environments.

Complementary advances in 3D-aware language models and AR-centric VLM pipelines tackle 3D understanding from the language side. Scene-LLM~\cite{fu2024scenellmextendinglanguagemodel} and 3D-LLM~\cite{hong3dllm} fuse 3D point clouds or meshes with language embeddings to answer questions about scanned environments, while hybrid systems such as XaiR~\cite{xair} project 3D scenes into 2D views and treat VLMs as reasoning layers on top of bespoke perception stacks. These efforts show that combining vision, language, and 3D structure improves spatial reasoning, but they typically assume high-quality, globally consistent 3D scans or pre-fused volumetric maps~\cite{10765411}, rely on dense segmentation or reconstruction~\cite{samr}, and are often evaluated offline rather than in low-latency, open-vocabulary AR sessions on commodity headsets.

\textbf{Gap.} Across these lines of work, no existing system simultaneously (1) supports open-vocabulary, language-guided queries over \emph{real} objects in live AR; (2) maintains an explicit, metric 3D representation that supports relational predicates such as \emph{behind}, \emph{next to}, and \emph{within reach}; and (3) operates within latency and robustness constraints compatible with interactive AR on commodity devices. Mesh-based augmentation~\cite{samr}, volumetric fusion pipelines~\cite{10765411}, and 3D-LLMs~\cite{fu2024scenellmextendinglanguagemodel,hong3dllm} provide rich 3D structure but depend on heavy scanning, reconstruction, or offline processing. Placement-oriented methods~\cite{octoPlus} and authoring agents~\cite{agentAR} focus on where to place virtual content or how to generate applications, not on answering relational questions about the existing physical scene. In-situ learning systems~\cite{huynhKelvar} yield persistent labels but neither expose a scene graph nor support open-vocabulary, relational querying across multiple objects.

\textbf{This paper introduces \emph{Words into World}, a task-adaptive AR agent that unifies the semantic breadth of MLLMs with the spatial grounding of depth-aware perception to support open-vocabulary, relational queries about real-world objects in live AR.} When a user issues a natural-language query, the system (1) captures a passthrough RGB frame on a Meta Quest~3, (2) uses an MLLM to propose context-aware, open-vocabulary object labels tailored to the query, (3) grounds these labels with an open-vocabulary detector, and (4) lifts 2D detections into metric 3D coordinates via pixel-to-depth raycasting on the headset's environment depth API. The resulting 3D anchors are aggregated into dynamic AR scene graphs that encode objects and a rich set of relationships, including spatial (e.g., \emph{on}, \emph{next to}, \emph{behind}, \emph{within reach}), temporal and sequential (\emph{before}/\emph{after}), causal, structural (\emph{part-of}), functional (\emph{used-for}), semantic (\emph{is-a}), dependency, interaction, and referential links between entities. This enables the agent to answer queries ranging from simple identification (\emph{``Which screwdriver is within reach on the desk?''}) to compound relational questions (\emph{``Which device that depends on the router is currently behind the monitor and was used after the laptop?''}). A task-adaptive controller selects appropriate spatial and relational tools---ray-based selection, local surface estimation, distance measurement, filtering, temporal and causal reasoning, and comparison---depending on the query structure and scene state.

\noindent\textbf{Contributions.} This work makes the following contributions:
\begin{enumerate}
\item \textbf{Task-adaptive AR agent for language-guided 3D and relational reasoning.} We present \emph{Words into World}, an AR agent that couples MLLM-based open-vocabulary semantics with a depth-based 2D-to-3D grounding module on a Meta Quest~3, enabling explicit reasoning about distances, visibility, and diverse object--object relations (spatial, temporal, causal, structural, functional, semantic, and beyond) in metric 3D space.
\item \textbf{Lightweight 3D scene graphs without full reconstruction.} We propose a depth-driven pipeline that lifts 2D detections into stable 3D anchors and organizes them into dynamic scene graphs, avoiding expensive per-object mesh reconstruction as in SAMR~\cite{samr} and pre-scanned volumetric fusion pipelines~\cite{10765411}, while still supporting rich relational predicates such as spatial (\emph{behind}, \emph{next to}, \emph{within reach}) and higher-level temporal, causal, structural, and functional relations.
\item \textbf{Language-conditioned relational reasoning in live AR.} We design a task-adaptive controller that parses natural-language queries, selects appropriate spatial and relational tools, and operates over the scene graph to answer open-vocabulary, multi-object relational queries in live AR sessions, complementing prior work on placement~\cite{octoPlus}, in-situ learning~\cite{huynhKelvar}, and language-guided AR agents~\cite{agentAR,xair}.
\item \textbf{GroundedAR-Bench benchmark and empirical evaluation.} We introduce \emph{GroundedAR-Bench}, a benchmark and suite of tasks for assessing language-conditioned spatial grounding, 3D localization accuracy, and relational reasoning across diverse indoor scenes, and we report quantitative and qualitative results comparing pipeline variants and analyzing trade-offs between spatial accuracy, semantic coverage, and latency.
\end{enumerate}

The remainder of this paper is organized as follows. Section~\ref{sec:related} situates our work within research on 3D scene understanding, open-vocabulary perception and placement, and language-guided AR interaction. Section~\ref{sec:method} describes the Words into World agent architecture, including the task-adaptive controller, 2D-to-3D grounding module, and relational reasoning over scene graphs. Section~\ref{sec:experiments} presents GroundedAR-Bench and our experimental setup, and Section~\ref{sec:results} reports quantitative and qualitative results comparing pipeline variants and analyzing system behavior across realistic AR scenarios. Section~\ref{sec:discussion} concludes with implications for future AR agents and directions for more deeply integrating 3D structure into multimodal language models.

\section{Related Work}\label{sec:related}

We organize related work into three areas: (1) vision--language models for 3D scene understanding, (2) open-vocabulary perception and object placement in AR, and (3) language-guided AR interaction and agentic systems.

\subsection{Vision--Language Models for 3D Scene Understanding}

Recent multimodal language models such as GPT-4o, Qwen-VL, and CLIP-based architectures have demonstrated strong performance in zero-shot recognition and open-vocabulary classification from images. However, as highlighted by SAMR~\cite{samr} and recent evaluations of VLMs in AR contexts, these models remain predominantly 2D: they infer semantics from images but lack explicit geometric representations, reliable depth estimation, or built-in mechanisms for reasoning over object--object relations and occlusion. When deployed as black-box agents in head-worn AR, they tend to misinterpret depth cues, confuse which object is in front or behind, and hallucinate spatial relationships when appearance alone is ambiguous.

To address this limitation, several lines of work have explored 3D-aware language models that incorporate point clouds, meshes, or voxel grids. Scene-LLM~\cite{fu2024scenellmextendinglanguagemodel} and 3D-LLM~\cite{hong3dllm}, for example, embed 3D scene representations and textual descriptions into a shared latent space, enabling question answering about object identities and relations within fully scanned environments. These methods, however, assume access to high-quality, globally consistent 3D reconstructions (e.g., from RGB-D scanning), which are seldom available in interactive AR where depth maps are sparse, noisy, or only locally valid. Moreover, their inference pipelines are typically offline or near-offline, limiting applicability to real-time, conversational AR.

Other work sidesteps direct 3D reasoning by projecting 3D scenes into multiview images and analyzing these images with 2D VLMs. While this improves coverage over the scene, it still relies on heuristic view selection and does not directly expose metric 3D quantities such as distances or reachability. Hybrid systems such as XaiR~\cite{xair} treat VLMs as high-level reasoning layers atop bespoke perception modules that handle object detection and pose estimation, but have largely been evaluated in static or scripted scenarios rather than live, open-vocabulary AR.

Multimodal 3D fusion systems for spatially aware AI push 3D understanding in AR further by explicitly fusing geometry and semantics into a volumetric representation. \emph{Multimodal 3D Fusion and In-Situ Learning for Spatially Aware AI}~\cite{10765411} constructs a multimodal 3D scene volume by integrating RGB, depth, and learned semantic features into a fused representation, enabling natural-language search and in-situ learning over reconstructed environments on head-worn displays. Their system demonstrates compelling object-centric retrieval and inventory management in AR workspaces; however, it relies on pre-scanned, densely fused 3D volumes and is optimized for operations over that static representation rather than for low-latency, on-demand reasoning in live, dynamically changing AR scenes.

Our work takes a complementary approach: rather than building a fully 3D language model or global volumetric map, we keep the MLLM in its native 2D+language regime and compensate for its spatial blind spots by coupling it with a depth-based 2D-to-3D grounding module on the headset. Words into World uses the MLLM for open-vocabulary candidate generation and intent parsing, and then lifts detections into metric 3D via pixel-to-depth raycasting, enabling explicit computation of distances, visibility, and relational predicates without requiring dense scene reconstruction or multi-pass volumetric fusion~\cite{10765411}. This design positions our agent between 2D-only VLM deployments and volumetric 3D fusion pipelines, focusing on interactive, query-conditioned 3D reasoning on commodity AR headsets.

\subsection{Open-Vocabulary Perception and Object Placement in AR}

Open-vocabulary detection (OVD) methods such as Grounding DINO, OWL-ViT, and CLIP-based detectors decouple object localization from closed-set classification, allowing arbitrary text queries to retrieve 2D bounding boxes for visual entities. These detectors have shown strong zero-shot performance across long-tail categories and are natural building blocks for AR scenarios where users may reference idiosyncratic or personalized objects. Prior AR systems primarily embed OVD for label detection or simple highlighting; they typically stop at 2D boxes on the image and do not integrate the results into a persistent 3D representation or scene graph, limiting their ability to support relational queries.

OCTO+~\cite{octoPlus} extends OVD into the domain of \emph{virtual object placement}. Given a single RGB view and a virtual object description (e.g., \emph{cupcake''}), OCTO+: (1) uses RAM++ and Grounding DINO to enumerate candidate supporting surfaces, (2) employs GPT-4 to reason about which surface is most semantically appropriate (e.g., a plate rather than the floor), and (3) uses G-SAM to pick a 2D placement point on the selected surface that is far from boundaries. This pipeline achieves state-of-the-art open-vocabulary placement performance on the PEARL benchmark, which quantifies how often placements fall inside valid regions and how deep'' they are within those regions. OCTO+ thus tackles the inverse of our problem: determining where to place \emph{new virtual content} in a way that is semantically natural, operating primarily in 2D and relying on a single raycast for 3D projection.

In contrast, Words into World focuses on \emph{grounding and relating existing real objects} rather than placing new virtual ones. Instead of selecting a single 2D point, we construct a set of 3D anchors and bounding volumes for many objects simultaneously by raycasting 2D detections into the headset's depth map. We then organize these anchors into a scene graph and support relational queries over that graph. Our pipeline is thus complementary to OCTO+: both build on open-vocabulary detectors and MLLM reasoning, but we target explicit, metric 3D scene understanding and relational retrieval in live AR rather than virtual content placement.

SAMR~\cite{samr} sits at the intersection of OVD and 3D grounding for MR. It segments 2D images with FastSAM, raycasts mask pixels into the HoloLens scene mesh to obtain object-level point clouds, and fits meshes for each object. These object meshes are then annotated with numeric identifiers and passed, along with the annotated image, to a VLM that answers questions about object identity, distance, and relations. SAMR demonstrates that augmenting VLMs with 3D structure substantially improves grounding, and reports high segmentation and recognition accuracy on cluttered desktops. However, mesh generation and VLM inference together yield end-to-end latencies on the order of 10--15 seconds per query, and the reliance on dense mask-to-mesh reconstruction limits scalability to larger scenes and more dynamic interaction.

Words into World adopts a lighter-weight 3D representation that trades dense object meshes for fast, depth-based 3D anchors. By operating directly on the Quest~3 environment depth API, we avoid per-object mesh fitting and instead estimate a small number of robust anchor points per detection, which are sufficient to compute reachability and topological relations. This design allows us to maintain an explicit scene graph and support interactive, query-conditioned reasoning while remaining within latency budgets compatible with real-time AR. In contrast to volumetric fusion approaches that operate over global TSDF volumes~\cite{10765411}, our representation is built incrementally from live views and is tightly coupled to the current user query and viewpoint.

\subsection{Language-Guided AR Interaction and Agentic Systems}

A parallel strand of work explores AR systems where large language models act as agents that interpret user intent, orchestrate perception modules, and generate AR content. Systems such as XaiR~\cite{xair} and ImaginateAR~\cite{imaginatear} show that LLMs can map natural-language requests to AR operations, retrieve relevant objects, and synthesize visual overlays. However, these systems often rely on coarse geometric proxies (e.g., object centers, 2D bounding boxes) or offline scene reconstructions, which constrain the complexity of spatial queries they can handle. They typically do not expose a structured 3D scene graph or support fine-grained relational predicates like \emph{behind}, \emph{closest to}, or \emph{within reach} grounded in metric space.

SAMR~\cite{samr} advances this line by integrating gaze, hand rays, and speech input for object selection, and by linking selected objects to a VLM capable of answering spatial questions. Yet interactions are primarily \emph{selection-based}: users explicitly point at one or more objects and then ask about them, and the system's understanding is scoped to those selected meshes. In contrast, Words into World supports \emph{open-ended, language-only} queries that jointly resolve referents and relations across the entire scene (e.g., \emph{``Which of the boxes on the shelf is behind the printer?''}) without requiring explicit preselection. Our task-adaptive controller parses query structure, triggers appropriate perception tools, and operates over the maintained scene graph to answer these questions.

Overall, our work extends language-guided AR agents along three dimensions: (1) we tightly couple MLLM-driven open-vocabulary semantics with depth-based 3D anchoring directly on commodity headsets; (2) we represent scenes as explicit graphs of objects and relations rather than as unstructured sets of boxes, meshes, or volumetric cells; and (3) we introduce GroundedAR-Bench to systematically evaluate language-conditioned spatial grounding and relational reasoning in realistic AR scenarios, providing a benchmark that complements placement-oriented evaluations such as PEARL and segmentation/recognition metrics used in prior MR systems, as well as volumetric fusion benchmarks for spatially aware AI~\cite{10765411}.

\vspace{-0.5em}
\section{Method}
\label{sec:method}

We implement \emph{Words into World} as a layered, agentic system that couples an on-device AR client on the Meta Quest~3 with a remote multimodal language--vision backend. Rather than a single monolithic pipeline, the system exposes a set of stateless tools, a persistent world model, and a small collection of cooperating agents that plan, call tools, and decide how to respond. This design lets us unify low-level 2D--to--3D grounding, rich object--object relationships, region-of-interest (ROI) highlighting, and geo-anchored memory within a single architecture that supports open-ended, conversational interaction.

We first formulate the task, then present the four-layer architecture (sensing \& actuation, tools, world model, and agent orchestration), and finally describe language-guided perception, 2D-to-3D grounding, relational reasoning, ROI highlighting, and location-aware persistence.

\subsection{Problem Formulation}

A user wearing a head-mounted AR device observes a physical scene. At time \(t\), the system has access to:
\begin{itemize}
    \item a monocular RGB frame \(I_t \in \mathbb{R}^{H \times W \times 3}\) from the passthrough camera,
    \item a per-pixel depth map \(D_t \in \mathbb{R}^{H \times W}\) aligned with \(I_t\),
    \item the current head pose and camera intrinsics,
    \item an audio stream that is transcribed into natural language utterances \(\{q_t\}\) via ASR, and
    \item optional geo-sensing and environment signals (GPS, compass, barometer, world-locked anchors).
\end{itemize}

The system maintains a world model consisting of:
\begin{equation*}
\mathcal{S}_t = \bigl\{ (o_i, \mathbf{p}_i, B_i, \ell_i, c_i) \bigr\}_{i=1}^N,
\quad
\mathcal{R}_t = \bigl\{ (o_i, o_j, r_{ij}, \gamma_{ij}) \bigr\},
\end{equation*}
where each object \(o_i\) is associated with a semantic label \(\ell_i\), a detector confidence \(c_i\), a metric 3D anchor \(\mathbf{p}_i \in \mathbb{R}^3\), and a 3D bounding volume \(B_i\), and where \(\mathcal{R}_t\) encodes typed relations \(r_{ij}\) between object pairs with confidence \(\gamma_{ij}\). Relations cover both geometric predicates (e.g., \emph{on}, \emph{behind}, \emph{within-reach}) and higher-level semantic/functional structure (detailed below).

Given a new utterance \(q_t\), the agent must:
\begin{enumerate*}[label=(\roman*)]
    \item update \(\mathcal{S}_t\) and \(\mathcal{R}_t\) via perception and grounding tools,
    \item interpret \(q_t\) as a task over this world model (Locate, Relate, Measure, Navigate, Recall, Share, \ldots),
    \item compute an answer grounded in 3D and/or geo space, and
    \item render AR overlays (bounding boxes, ROIs, relation edges) and multimodal feedback (audio, haptics) that help the user act on that answer.
\end{enumerate*}

\subsection{Layered Agent Architecture}

Figure~\ref{fig:system-architecture}  summarizes our four-layer architecture: (1)~a \emph{Sensing \& Actuation Layer} on-device, (2)~a \emph{Agent Layer} of callable capabilities, (3)~a shared \emph{World Model Layer}, and (4)~an \emph{ Orchestration Layer} that plans and decides how to respond.

\subsubsection{Sensing \& Actuation Layer (on-device)}

The sensing \& actuation layer runs entirely on the Meta Quest~3 and provides raw I/O, but performs no high-level reasoning.

\paragraph{Inputs}
\begin{itemize}
    \item \textbf{Vision \& depth.} Passthrough RGB frames \(I_t\), environment depth maps, and (when available) scene meshes.
    \item \textbf{Head \& hand pose.} Headset tracking and controller/hand-joint poses for ray-based selection and gestures.
    \item \textbf{Audio in.} Microphone input for continuous ASR, enabling hands-free, conversational queries.
    \item \textbf{Geo \& environment.} GPS, compass/magnetometer, barometer, and world-locked anchors (OVRSpatialAnchors) that tie AR coordinates to physical locations.
    \item \textbf{Gestures.} Pinch, grab, and point gestures, interpreted as high-level control signals (confirm, disambiguate, select).
\end{itemize}

\paragraph{Outputs}
\begin{itemize}
    \item \textbf{AR visuals.} 3D bounding boxes, pins at object anchors, relation edges, ROI highlights, and navigation ``breadcrumbs''.
    \item \textbf{Haptics.} Distance-graded vibration and directional buzz patterns indicating proximity and orientation of targets.
    \item \textbf{Spatial audio.} Directional beacons from object/waypoint locations and spoken guidance.
    \item \textbf{Gesture-driven actions.} Mapping of gestures (e.g., pinch) to agent-level intents (e.g., confirm a choice, ``grab'' a virtual proxy).
\end{itemize}

\subsubsection{Tool Layer}

From the agent's perspective, tools are stateless functions: each call takes a structured input, returns a result, and leaves long-term state to the world model.

\paragraph{Language input \& planning tools}
A \emph{Voice-to-Action Parser} converts the ASR text stream into segmented utterances, intent hypotheses, and coarse plan sketches (e.g., ``need proximity filter; raycast sphere around lamp''). A \emph{Query Clarification Loop} evaluates ambiguity given the current world model and, when confidence is low or multiple objects satisfy a description, generates clarification prompts (``Do you mean the red or blue book?'') and integrates user responses into the plan and scene graph.

\paragraph{Perception \& 3D grounding tools}
A \emph{Frame Capture} tool returns synchronized RGB, depth, and head pose. An \emph{Open-Vocabulary Label Proposer} (MLLM) takes \(q_t\) and \(I_t\) and yields candidate object labels, rough regions, and textual descriptions. A \emph{2D Detector / Grounder} (e.g., Grounding DINO) consumes these labels and returns 2D bounding boxes and confidences. A \emph{Raycast \& 3D Lifting} tool performs multi-ray sampling into the depth map or scene mesh to obtain 3D anchors, extents, and support planes. Finally, a \emph{Scene Graph Builder} fuses these results into a local relational graph over objects, surfaces, and the user.

\paragraph{Geo-anchoring \& navigation tools}
A \emph{Session Geo-Anchor Manager} uses GPS, compass, and the AR origin to maintain a session geo anchor (lat, lon, alt, orientation), stored as an OVRSpatialAnchor with geo metadata. An \emph{Object Geo-Projection} tool maps object anchors into geo coordinates and, optionally, addresses via reverse geocoding. A \emph{Geo Graph \& Pathfinding} module maintains a graph over user positions, object geo nodes, and landmarks and supports shortest-path queries and geo-fence checks. \emph{Geo-Share Tools} package geo + semantics into links/QR codes and resolve incoming tokens into AR anchors on other devices.

\paragraph{Memory \& adaptation tools}
\emph{Session-Long Spatial Memory} uses world-locked anchors and periodic graph updates to track how object positions and relations change within a session. \emph{Cross-Session Geo Memory} stores geo-tagged objects, notes, and overlays, enabling queries such as ``Relocate my bag from yesterday.'' A \emph{History \& Personalization} tool logs queries, clarifications, and outcomes for learning user preferences and recurring patterns.

\paragraph{Interaction \& output tools}
A \emph{Visual Overlay Manager} renders bounding boxes, pins, relation edges, ROI highlights, and geo badges. A \emph{Multimodal Feedback Suite} coordinates haptics (distance-graded vibration, directional buzz), spatial audio (beacons, spoken guidance), and ROIs. \emph{Navigation \& Proximity Alerts} emit notifications (e.g., ``Your keys are 20\,m away at home coords---want a map?'') and drive turn-by-turn AR breadcrumbs.

\subsubsection{World Model Layer}

All tools and agents read and write to a shared world model that acts as a blackboard.

\paragraph{Local scene graph}
The local scene graph encodes:
\begin{itemize}
    \item \textbf{Nodes} for objects, surfaces, and the user, each with 3D pose, size, appearance label, detector confidence, and last-seen timestamp;
    \item \textbf{Edges} for spatial relations such as \emph{on}, \emph{under}, \emph{next-to}, \emph{behind}, \emph{within-reach}, \emph{closest-to}, and \emph{blocking}, as well as occlusion, reachability, and proximity.
\end{itemize}

\paragraph{Geo graph}
The geo graph contains:
\begin{itemize}
    \item \textbf{Nodes} for session geo anchors, object geo nodes (lat, lon, alt, accuracy, address), and landmarks (desk, door, ``home base''),
    \item \textbf{Edges} for spatial adjacency, navigable connections (corridors, stairs), and ``stored at'' relationships (e.g., keys stored at HomeEntranceAnchor).
\end{itemize}

\paragraph{Memory state}
Session memory stores a history of scene graphs and object trajectories (e.g., ``mug moved from desk to shelf''), while cross-session memory archives geo-graphs per location, user-authored pins, and ROIs. A user model records preferences (e.g., haptic-heavy feedback) and typical objects of interest.

\paragraph{Uncertainty \& health}
We maintain confidence scores per node/edge, sensor health flags (e.g., GPS weak, depth noisy), and a revision history to support re-anchoring when drift or conflicts are detected.

\subsubsection{Agent / Orchestration Layer}

The orchestration layer is the ``mind'' of the system: a Supervisor Agent coordinates four sub-agents that share the world model and call tools.

\paragraph{Dialogue \& planning agent}
The dialogue agent listens to the Voice-to-Action Parser, infers task types (Locate, Relate, Measure, Navigate, Recall, Share, \ldots), and constructs task graphs in a ReAct-style format (Thoughts and Actions). It consults the Query Clarification Loop when ambiguity is high and produces high-level plans such as:
\emph{``Refresh scene graph $\rightarrow$ filter objects on desk $\rightarrow$ choose mug closest to laptop $\rightarrow$ project to geo $\rightarrow$ guide user.''}

\paragraph{Perception \& grounding agent}
The perception agent executes the ``refresh scene'' parts of plans: it decides when to capture new frames versus reuse cached ones, calls label proposer, detector, 3D lifting, and the Scene Graph Builder, and performs self-checking (e.g., asking the user to point if detection or depth are unreliable).

\paragraph{Navigation \& geo agent}
The navigation agent handles queries involving geo-level reasoning (``Where exactly is my phone?'', ``Guide me from here to the emergency stop button, avoid stairs''), using geo anchoring, pathfinding, and the feedback suite to generate AR breadcrumbs, haptic gradients, and spatial audio beacons. It also manages geo-fence logic (e.g., ``Grab your badge before you leave the office'').

\paragraph{Memory \& sharing agent}
The memory agent manages session-long and cross-session memory, surfaces insights such as ``You moved the mug again; it's now on the shelf,'' and resolves recall queries such as ``Hide the mug again, but ignore the broken one.'' It also controls geo-sharing, creating and interpreting shareable links/QR codes and synchronizing anchors across devices.

A feedback policy module decides \emph{how} to answer (visual vs.\ audio vs.\ haptic, brief vs.\ step-by-step) based on user preferences and context (e.g., eyes-busy vs.\ focused).

\subsection{Language-Guided Perception and Scene Graph Construction}

Within this architecture, language-guided perception is implemented as a composition of tools orchestrated by the Dialogue and Perception agents. Conceptually, it mirrors our original pipeline but is now embedded in a recurrent plan--perceive--update loop.

\subsubsection{Intent parsing and scene analysis}

Given an utterance \(q_t\), the dialogue agent uses the Voice-to-Action Parser to infer an initial task type and plan sketch. For perception-heavy tasks, it requests a snapshot via Frame Capture, obtaining \((I_t, D_t, \text{pose}_t)\), and calls the Open-Vocabulary Label Proposer with \((q_t, I_t)\). The label proposer returns:
\begin{itemize}
    \item a global natural language description of the scene,
    \item a set of query-relevant candidate objects with labels and textual descriptions,
    \item optional hypotheses about their relations (e.g., ``the mug is on the desk and in front of the laptop'').
\end{itemize}
These labels are converted into text prompts for the 2D Detector / Grounder, which yields 2D bounding boxes and confidences.

\subsubsection{2D-to-3D grounding via depth raycasting}

The Raycast \& 3D Lifting tool converts 2D detections into 3D anchors and volumes. For each detected object \(o_i\) with bounding box \(b_i\), we:
\begin{enumerate}
    \item Sample several pixels inside \(b_i\) (center, corners, and points along diagonals) to hedge against local noise.
    \item For each pixel \(\mathbf{u}\), construct a 3D ray using camera intrinsics and the current head pose, and intersect this ray with the depth map \(D_t\) (or scene mesh) to obtain candidate hit points \(\hat{\mathbf{p}}\).
    \item Aggregate valid hits, compute a robust depth estimate (e.g., median), and reject outliers.
    \item Estimate extents and surface-aligned orientation from the spatial spread and local normals.
\end{enumerate}
The resulting anchor \(\mathbf{p}_i\) and bounding volume \(B_i\) are written to the local scene graph as a node with label \(\ell_i\), confidence \(c_i\), and last-seen timestamp.

\subsubsection{Relational reasoning and graph types}

The Scene Graph Builder then instantiates edges \(\mathcal{R}_t\) that capture both spatial and higher-level relationships. We distinguish two complementary families:

\paragraph{Spatial and topological relations}
Using the 3D geometry of anchors and volumes, we infer relations such as \emph{on}, \emph{under}, \emph{next-to}, \emph{behind}, \emph{within-reach}, \emph{closest-to}, and \emph{blocking}. These edges are computed algorithmically from depth, distances, relative orientations, and simple reachability heuristics (e.g., arm-length thresholds).

\paragraph{Semantic, functional, and temporal relations}
To capture richer structure in complex scenes (e.g., workstations, tool setups), we further classify edges into interpretable types, prompted from the MLLM and grounded whenever possible in geometry and object roles:
\begin{itemize}
    \item \textbf{Sequential}: \(A \rightarrow B\) (A happens before B),
    \item \textbf{Causal}: \(A \rightarrow B\) (A causes or enables B),
    \item \textbf{Structural}: \(A \leftrightarrow B\) (A is a part or sub-component of B),
    \item \textbf{Functional}: \(A \rightarrow B\) (A is a function or tool for B),
    \item \textbf{Semantic}: \(A \rightarrow B\) (A is a semantic subtype of B),
    \item \textbf{Dependence}: \(A \rightarrow B\) (A exist or presents before B),
    \item \textbf{Interaction}: \(A \leftrightarrow B\) (A physically interacts with B),
    \item \textbf{Referential}: \(A \rightarrow B\) (A refers to or points to B).
\end{itemize}

Practically, the Perception agent passes the current scene graph skeleton (nodes and spatial relations) plus \(q_t\) to the MLLM, which proposes candidate edges with types and justifications. The Scene Graph Builder then reconciles these proposals with hard geometric constraints (e.g., enforcing that ``on'' implies vertical support, rejecting causal claims inconsistent with known object affordances) and assigns confidences \(\gamma_{ij}\) to each edge.

We formalize this fusion of geometric and semantic evidence as a hybrid relation inference module, shown in Algorithm~\ref{alg:relation_inference}. The module takes the current object set \(\mathcal{S}_t\), user position \(\mathbf{p}_{\text{user}}\), and LLM-proposed relations \(\mathcal{E}^{\mathrm{LLM}}\), and produces the final relation set \(\mathcal{R}_t\) used throughout the world model.

\begin{algorithm*}[t]
\caption{Hybrid Geometric--Semantic Relation Inference}
\label{alg:relation_inference}
\begin{algorithmic}[1]
\REQUIRE Object set $\mathcal{S}_t = \{(o_i, \mathbf{p}_i, B_i, \ell_i, c_i)\}_{i=1}^N$, user position $\mathbf{p}_{\text{user}}$, LLM proposals $\mathcal{E}^{\mathrm{LLM}} = \{(i,j,r, s^{\mathrm{LLM}}_{ij}(r))\}$, hyperparameters $\alpha \in [0,1]$, $\tau \in (0,1)$
\ENSURE Relation set $\mathcal{R}_t = \{(o_i, o_j, r_{ij}, \gamma_{ij})\}$

\STATE Define spatial relation types $\mathcal{R}_{\mathrm{sp}} = \{\text{on}, \text{under}, \text{next-to}, \text{behind}, \text{within-reach}, \text{closest-to}, \text{blocking}\}$
\STATE Define high-level relation types $\mathcal{R}_{\mathrm{hl}} = \{\text{Sequential}, \text{Causal}, \text{Structural}, \text{Functional}, \text{Semantic}, \text{Dependence}, \text{Interaction}, \text{Referential}\}$
\STATE Initialize $\mathcal{R}_t \gets \varnothing$
\STATE Precompute user distances $d_i^{\text{user}} = \|\mathbf{p}_i - \mathbf{p}_{\text{user}}\|_2$ for all $i$

\FOR{each ordered pair of objects $(o_i, o_j)$ with $i \neq j$}
    \STATE Compute pairwise distance $d_{ij} = \|\mathbf{p}_i - \mathbf{p}_j\|_2$
    \STATE Let $\Delta z_{ij} = (\mathbf{p}_j)_z - (\mathbf{p}_i)_z$ and horizontal offset $h_{ij} = \|[(\mathbf{p}_i)_x, (\mathbf{p}_i)_y] - [(\mathbf{p}_j)_x, (\mathbf{p}_j)_y]\|_2$
    \STATE Initialize geometric scores $s^{\mathrm{geo}}_{ij}(r) \gets 0$ for all $r \in \mathcal{R}_{\mathrm{sp}} \cup \mathcal{R}_{\mathrm{hl}}$
    
    \STATE \COMMENT{Example geometric predicates (others follow analogous heuristics)}
    \STATE $s^{\mathrm{geo}}_{ij}(\text{within-reach}) \gets \mathbb{I}[d_i^{\text{user}} \leq R_{\text{reach}}]$
    \STATE $s^{\mathrm{geo}}_{ij}(\text{on}) \gets \mathbb{I}[|\Delta z_{ij}| \leq \varepsilon_z \land h_{ij} \leq \varepsilon_h \land \text{support}(B_i, B_j)]$
    \STATE $s^{\mathrm{geo}}_{ij}(\text{behind}) \gets \mathbb{I}[(\mathbf{p}_i)_z > (\mathbf{p}_j)_z + \varepsilon_{\text{depth}}]$
    \STATE $s^{\mathrm{geo}}_{ij}(\text{closest-to}) \gets \mathbb{I}[d_{ij} = \min_{k \neq j} d_{kj}]$
    
    \STATE \COMMENT{For high-level types, geometry may be uninformative: $s^{\mathrm{geo}}_{ij}(r) = 0$ for $r \in \mathcal{R}_{\mathrm{hl}}$}
    \STATE Retrieve LLM scores $s^{\mathrm{LLM}}_{ij}(r)$ for any $(i,j,r, s^{\mathrm{LLM}}_{ij}(r)) \in \mathcal{E}^{\mathrm{LLM}}$; otherwise set $s^{\mathrm{LLM}}_{ij}(r) \gets 0$
    
    \STATE \COMMENT{Fuse geometric and LLM scores for all relation types}
    \FOR{each relation type $r \in \mathcal{R}_{\mathrm{sp}} \cup \mathcal{R}_{\mathrm{hl}}$}
        \STATE Compute fused score $\gamma_{ij}(r) \gets \sigma\bigl(\alpha\, s^{\mathrm{LLM}}_{ij}(r) + (1-\alpha)\, s^{\mathrm{geo}}_{ij}(r)\bigr)$, where $\sigma(x) = 1/(1 + e^{-x})$
    \ENDFOR
    
    \STATE Select best relation type: $r_{ij}^\star \gets \arg\max_{r \in \mathcal{R}_{\mathrm{sp}} \cup \mathcal{R}_{\mathrm{hl}}} \gamma_{ij}(r)$, \quad $\gamma_{ij}^\star \gets \max_{r} \gamma_{ij}(r)$
    \IF{$\gamma_{ij}^\star \geq \tau$}
        \STATE Add edge $(o_i, o_j, r_{ij}^\star, \gamma_{ij}^\star)$ to $\mathcal{R}_t$
    \ENDIF
\ENDFOR

\STATE \COMMENT{Optional: enforce symmetry for bi-directional types}
\FOR{each $(o_i, o_j, r_{ij}, \gamma_{ij}) \in \mathcal{R}_t$ with $r_{ij} \in \{\text{Structural}, \text{Interaction}\}$}
    \IF{$(o_j, o_i, r_{ji}, \gamma_{ji}) \notin \mathcal{R}_t$}
        \STATE Add $(o_j, o_i, r_{ij}, \gamma_{ij})$ to $\mathcal{R}_t$
    \ENDIF
\ENDFOR

\RETURN $\mathcal{R}_t$
\end{algorithmic}
\end{algorithm*}

The resulting graph \(G = (V, E)\) is thus stored as a set of triples \((o_i, o_j, r_{ij}, \gamma_{ij})\), enabling queries that mix geometric and semantic structure (e.g., ``Which tools functionally depend on the soldering station and are within reach on the desk?'').

\subsection{ROI Highlighting, Relationship Visualization, and Multimodal Feedback}

ROI highlighting and relationship visualization are implemented by the Visual Overlay Manager but grounded in the world model.

\subsubsection{ROI highlighting}

For each node \(o_i\) selected as relevant to the current plan (e.g., all objects matching the query, or the subgraph answering a relational question), the overlay manager:
\begin{itemize}
    \item projects its 3D anchor \(\mathbf{p}_i\) to a 2D screen coordinate \(\hat{\mathbf{u}}_i\),
    \item instantiates a world-locked \emph{pin} at \(\mathbf{p}_i\), and
    \item draws a semi-transparent ROI around \(\hat{\mathbf{u}}_i\), either filling the bounding box or using a radial falloff mask.
\end{itemize}

ROI intensity and color are modulated by relevance, relation type, and confidence: for example, objects that \emph{functionally} depend on the laptop might share one hue, while objects in a \emph{structural} relation (components of a device) use another. This turns the scene into a spatially anchored, query-specific heatmap that remains visually grounded in the passthrough view.

\subsubsection{Relationship mapping and scene graph overlays}

For each edge \((o_i, o_j, r_{ij}, \gamma_{ij})\) selected by the plan, the overlay manager renders a 3D link between \(\mathbf{p}_i\) and \(\mathbf{p}_j\) with line style encoding \(r_{ij}\) (e.g., dashed for temporal/Sequential, double-line for Structural, arrowheads for Causal/Dependence, bidirectional styling for Structural/Interaction). Because real scenes can be densely connected, the Dialogue agent selects a subgraph relevant to the current query (e.g., all objects \emph{on} the desk and \emph{functionally} connected to the laptop) rather than visualizing the entire graph at once.

Together, ROIs and relation edges turn the abstract scene graph into an interactive AR visualization: users can not only locate individual objects but also \emph{see} how components, tools, and surfaces relate in 3D.

\subsection{Location-Aware Persistence and Geo-Level Reasoning}

Location-aware persistence and navigation are handled jointly by the Navigation \& Geo agent and the Memory \& Sharing agent.

\subsubsection{Session-long spatial memory}

Within a session, the Session-Long Spatial Memory tool periodically snapshots \((\mathcal{S}_t, \mathcal{R}_t)\) and associates them with world-locked anchors. When the user looks away and returns, pins, ROIs, and relation edges remain stable. The memory agent can also surface changes (e.g., ``You moved the mug---it's now on the shelf'') by comparing successive scene graphs and updating edge types (e.g., new Spatial and Dependence relations).

\subsubsection{Cross-session geo memory}

For cross-session recall, the Cross-Session Geo Memory tool uses the Session Geo-Anchor Manager and Object Geo-Projection to store tuples \((g_t, \mathcal{S}_t, \mathcal{R}_t)\), where \(g_t\) encodes geo coordinates and heading. When the user later asks ``Where did I leave my mug yesterday?'', the memory agent retrieves the most relevant mug node and geo anchor from history, the navigation agent computes a path in the geo graph, and the feedback suite renders AR breadcrumbs, haptic gradients, and spatial audio beacons. Upon arrival, the perception agent refreshes the local scene graph and reinstates precise 3D boxes, ROIs, and relation edges around the mug.

Commands such as ``Hide the mug again, but ignore the broken one'' are executed by reusing prior hiding-location regions from memory, filtering out objects with Semantic or Structural attributes indicating ``broken,'' and updating both spatial and geo memory so future ``hide again'' requests reuse the new state.

\subsection{Agentic Inference Cycle and Implementation Details}

Algorithm~\ref{alg:agent_cycle} summarizes a typical agentic inference cycle for a spatial or relational query.

\begin{algorithm}[t]
\caption{Agentic Inference Cycle for a Spatial/Relational Query}
\label{alg:agent_cycle}
\begin{algorithmic}[1]
\REQUIRE ASR stream, current world model $(\mathcal{S}_t, \mathcal{R}_t, \mathcal{G}_t)$
\ENSURE Grounded answer, AR overlays, and feedback
\STATE Dialogue agent receives utterance $q_t$ from Voice-to-Action Parser
\STATE Infer task type (Locate / Relate / Measure / Navigate / Recall / Share)
\IF{world model sufficient and confidence high}
    \STATE Answer directly from $(\mathcal{S}_t, \mathcal{R}_t, \mathcal{G}_t)$
\ELSE
    \STATE Perception agent calls Frame Capture $\rightarrow (I_t, D_t, \text{pose}_t)$
    \STATE Call Open-Vocabulary Label Proposer on $(q_t, I_t)$
    \STATE Call 2D Detector / Grounder with labels $\rightarrow$ 2D boxes
    \STATE Call Raycast \& 3D Lifting with $(\text{boxes}, D_t, \text{pose}_t)$
    \STATE Update local scene graph via Scene Graph Builder, including relation types
\ENDIF
\STATE Dialogue agent checks ambiguity via Query Clarification Loop; possibly refines targets
\STATE Compute answer over $(\mathcal{S}_t, \mathcal{R}_t, \mathcal{G}_t)$
\STATE Feedback policy selects output modalities and which subgraph to visualize
\STATE Visual Overlay Manager renders boxes, ROIs, and relation edges
\STATE Navigation \& Geo agent optionally activates breadcrumbs / proximity alerts
\STATE Memory \& Sharing agent logs $(q_t, \mathcal{S}_t, \mathcal{R}_t, \mathcal{G}_t)$ to session and geo memory
\end{algorithmic}
\end{algorithm}

The client is implemented in Unity on Meta Quest~3, using the platform's passthrough and environment depth APIs, OVRSpatialAnchors for world-locked anchors, and built-in haptics and spatial audio for feedback. Gestures (pinch, grab, point) are detected via the Quest hand-tracking SDK. The backend runs on a local machine or edge server and exposes REST endpoints hosting the multimodal language model and open-vocabulary detector. All 3D raycasting, anchor construction, ROI rendering, and scene-graph visualization are computed on-device; semantic inference and open-vocabulary detection are offloaded to the backend.

End-to-end latency is dominated by MLLM and OVD calls, but the layered, agentic design keeps the interaction model stable while allowing components to be optimized or replaced (e.g., with smaller on-device models) as hardware improves. Crucially, the same architecture supports the full spectrum of behaviors described in this work: metric 3D grounding, rich relational scene graphs, ROI-based attention guidance, and geo-anchored memory and sharing.

\section{Experiments}
\label{sec:experiments}

We evaluate \emph{Words into World} along four dimensions: (1)~3D localization accuracy of language-referred objects, (2)~correctness of inferred object--object relationships and scene graphs, (3)~end-to-end performance on language-guided spatial retrieval tasks in realistic AR scenarios, and (4)~efficiency and robustness of the task-adaptive pipeline. All experiments are conducted on the Meta Quest~3 using our prototype implementation and the \emph{GroundedAR-Bench} dataset introduced below.

\subsection{Experimental Setup}

\subsubsection{Hardware and software}
The client runs on a Meta Quest~3 headset using passthrough RGB, the environment depth API, and head pose tracking. The backend runs on a local machine with a recent GPU, hosting the multimodal language model and Grounding DINO detector exposed through a lightweight HTTP service. Unless otherwise noted, we use the same model configuration as in Section~\ref{sec:method} and a detection confidence threshold of 0.35.

\subsubsection{GroundedAR-Bench}
To evaluate language-conditioned spatial grounding in realistic environments, we construct \emph{GroundedAR-Bench}, a collection of Quest-captured scenes covering three usage contexts:
\begin{itemize}
    \item \textbf{Industrial-like}: workbenches with tools, parts, and small assemblies;
    \item \textbf{Assistive}: kitchen and living-room areas with everyday objects (tableware, containers, medication, furniture);
    \item \textbf{General/desk}: office desks with laptops, monitors, peripherals, books, and personal items.
\end{itemize}

For each context we record multiple \emph{tidy} scenes (few objects, limited occlusion) and \emph{cluttered} scenes (many objects, occlusion, overlapping layouts). For every scene we store a synchronized RGB frame, depth map, and head pose, and annotate:
\begin{itemize}
    \item object instances with category labels and 2D bounding boxes;
    \item metric 3D object centers in the headset coordinate frame (obtained via in-situ alignment with a calibrated AR reference marker);
    \item a scene graph with directed edges and relation types drawn from the taxonomy in Section~\ref{sec:method};
    \item a set of natural language queries spanning identification, relational reasoning, and tool-based tasks (measurement, filtering, selection).
\end{itemize}

All queries are authored by two researchers and verified by a third annotator to ensure that the correct answer is unambiguous given the visible scene.

\subsubsection{Baselines}
Where appropriate, we compare \emph{Words into World} to two baselines:
\begin{itemize}
    \item \textbf{2D VLM baseline}: the multimodal LLM directly answers queries from the RGB image without any explicit 3D grounding; for localization tasks we project the 2D box center onto a default horizontal plane.
    \item \textbf{No-depth baseline}: our full pipeline without depth-based raycasting; 2D detections are lifted to 3D by intersecting image rays with a fitted support plane (e.g., desk or floor) estimated from sparse points.
\end{itemize}

These baselines represent typical 2D-centric and geometry-poor AR integrations of VLMs.

\subsection{Task 1: 3D Localization Accuracy}

The first experiment measures how accurately the system localizes language-referred objects in 3D.

\subsubsection{Protocol}
For each scene in GroundedAR-Bench we select a subset of annotated objects and issue simple identification queries of the form ``Where is the \emph{<object>}?'' or ``Highlight all \emph{<category>} objects.'' The agent returns a 3D anchor and bounding volume for each detected instance. We record, for each ground-truth object, the nearest predicted anchor of the same category (if any).

\subsubsection{Metrics}
We report:
\begin{itemize}
    \item \textbf{3D position error} (cm): Euclidean distance between predicted and ground-truth centers.
    \item \textbf{Angular error} (deg): angle between predicted and ground-truth direction vectors from the headset origin.
    \item \textbf{Success@$\tau$}: percentage of objects with 3D error below thresholds (e.g., $\tau \in \{10,20\}$\,cm).
    \item \textbf{2D IoU}: overlap between the projected predicted box and ground-truth 2D bounding box.
\end{itemize}

We compute metrics per context (industrial, assistive, general) and per scene difficulty (tidy vs.\ cluttered), and compare our method to both baselines. This experiment isolates the contribution of depth-based raycasting and robust multi-point sampling in the 2D-to-3D grounding pipeline.

\subsection{Task 2: Relation Grounding and Scene Graph Accuracy}

The second experiment evaluates the correctness of the object--object relationships inferred by the relationship reasoning module.

\subsubsection{Protocol}
For each scene we treat annotated objects as graph nodes and use the ground-truth scene graph as reference. We run the full pipeline with the coordinator instructed to enable relationship reasoning. The backend returns a set of predicted triples $(o_i,o_j,r_{ij})$, where $r_{ij}$ falls into one of the relation types defined in Section~\ref{sec:method}. We match predicted and ground-truth edges by object identity and relation type.

In addition, we evaluate query-driven relational retrieval using templates such as:
\begin{itemize}
    \item ``Which object is on top of the \emph{<object>}?'',
    \item ``Highlight everything that is next to the \emph{<object>},''\!
    \item ``Which devices are connected to the laptop on the desk?''
\end{itemize}
A query is considered successful if the set of highlighted objects matches the objects connected by the corresponding ground-truth relation.

\subsubsection{Metrics}
We report:
\begin{itemize}
    \item \textbf{Edge precision/recall/F1}: treating each directed, typed edge as a binary classification.
    \item \textbf{Relation-type accuracy}: fraction of correctly predicted relation labels for edges whose endpoints are correct.
    \item \textbf{Relational query success}: proportion of relational queries where the retrieved objects exactly match the ground-truth set.
\end{itemize}

These measures quantify how well the agent's inferred scene graph aligns with the physical scene and whether it supports reliable relational reasoning in AR.

\subsection{Task 3: Language-Guided Spatial Retrieval}

The third experiment assesses end-to-end performance on language-guided spatial retrieval tasks that combine detection, grounding, and reasoning.

\subsubsection{Query categories}
We group GroundedAR-Bench queries into four categories:
\begin{itemize}
    \item \textbf{Identification}: locate and highlight specific objects (``Where is my mug?'').
    \item \textbf{Relational}: retrieve objects based on spatial relations (``What is closest to the laptop and touching the desk surface?'').
    \item \textbf{Tool-based}: invoke measurement or filtering operations (``Measure the distance from the robot arm to the emergency stop button''; ``Select all objects within 20\,cm of the table edge.'').
    \item \textbf{Assistive}: support situational awareness (``Highlight the nearest door''; ``Show potential tripping hazards.'').
\end{itemize}

\subsubsection{Protocol and metrics}
For each query we log the set of objects highlighted in AR (and, when applicable, numeric outputs such as distances). Human annotators judge correctness with respect to the annotated scene; measurement queries are evaluated with an absolute error threshold. We report:
\begin{itemize}
    \item \textbf{Task success rate}: proportion of queries in each category for which the returned result matches the ground truth.
    \item \textbf{Distance error}: mean and median absolute error for measurement queries.
    \item \textbf{Category breakdown}: success rates per environment type and per query category.
\end{itemize}

We compare the full \emph{Words into World} agent against the 2D VLM baseline to quantify the benefit of explicit 3D grounding for language-conditioned interaction.

\subsection{Task 4: Ablation and Latency Analysis}

Finally, we analyze the efficiency and design choices of the pipeline.

\subsubsection{Ablation settings}
We consider the following variants:
\begin{itemize}
    \item \textbf{Full agent}: task-adaptive coordinator, relationship reasoning, and depth-based 2D-to-3D grounding.
    \item \textbf{No coordinator}: always execute the most expensive pipeline (full relationship reasoning and tools) regardless of query.
    \item \textbf{No depth}: identical to the full agent but with 3D anchors obtained by intersecting rays with a planar fit instead of querying the depth map.
\end{itemize}

\subsubsection{Metrics}
For each variant and query category we measure:
\begin{itemize}
    \item \textbf{End-to-end latency}: time from user trigger to AR overlay update, decomposed into capture/encoding, network transfer, MLLM processing, detection, and client-side grounding.
    \item \textbf{Task performance}: task success rates on Tasks 1--3 under each variant.
\end{itemize}

This experiment highlights trade-offs between semantic richness, spatial accuracy, and responsiveness, and demonstrates that the task-adaptive controller reduces latency while preserving accuracy, and that depth-guided grounding is essential for reliable 3D spatial retrieval.

\begin{table*}[t]
\centering
\caption{Overview of evaluation tasks: research questions, scene types, metrics, and baselines.}
\label{tab:exp-overview}
\small
\begin{tabular}{@{}p{1.8cm}p{3.2cm}p{3.8cm}p{3.5cm}p{2.5cm}@{}}
\toprule
\textbf{Task} & \textbf{Primary Research Question} & \textbf{Scenes / Queries} & \textbf{Metrics} & \textbf{Baselines} \\
\midrule

\textbf{T1: 3D Localization} & 
How accurately does the agent localize language-referred objects in 3D space? & 
Industrial, assistive, and desk scenes; tidy \& cluttered variants; simple identification queries (e.g., ``Where is the mug?''). & 
3D position error (cm); angular error (deg); Success@r; 2D IoU between projected box and ground truth. & 
2D VLM (no explicit 3D); no-depth variant (planar projection). \\

\midrule

\textbf{T2: Relation Grounding} & 
Can the agent correctly infer and ground object--object relations in a scene graph? & 
Scenes with annotated relations (on-top-of, next-to, connected-to, part-of, etc.); relational queries (e.g., ``What is next to the laptop?''). & 
Edge precision/recall/F1; relation-type accuracy; relational query success rate. & 
2D VLM textual reasoning only; detector-only graph built from geometric heuristics. \\

\midrule

\textbf{T3: Language-Guided Spatial Retrieval} & 
How well does the full pipeline support language-guided retrieval and tool-style queries in AR? & 
All scene types; mixed queries: identification, relational, measurement, filtering, assistive (e.g., ``Select objects within 20\,cm of the table edge.''). & 
Task success rate per query category; distance error for measurement tasks; per-context breakdown (industrial / assistive / desk). & 
2D VLM baseline; no-depth variant. \\

\midrule

\textbf{T4: Ablation \& Latency} & 
What are the trade-offs between semantic richness, spatial accuracy, and responsiveness? & 
Representative subset of scenes and queries from T1--T3. & 
End-to-end latency (capture, network, MLLM, detection, grounding); task success rates under ablations. & 
Full agent; no-coordinator (always full pipeline); no-depth. \\

\bottomrule
\end{tabular}
\end{table*}

\begin{table*}[t]
\centering
\caption{Qualitative comparison of depth-based versus scene-mesh raycasting for 3D anchoring on Quest~3. Our experiments use depth as the primary signal and treat the scene mesh as an optional prior or fallback.}
\label{tab:depth-vs-mesh}
\small
\begin{tabular}{@{}p{2.8cm}p{6cm}p{6cm}@{}}
\toprule
\textbf{Aspect} & \textbf{Depth-based Raycasting} & \textbf{Scene-mesh Raycasting} \\
\midrule

\textbf{Local metric accuracy} & 
Uses per-pixel depth aligned with the current RGB frame; typically more accurate along the view ray for small, nearby objects (mugs, tools, peripherals). & 
Operates on a fused, decimated mesh; surfaces are smoothed and planarized, which can shift hits slightly behind the true surface and blur small objects. \\

\midrule

\textbf{Spatial resolution} & 
Full image resolution (one depth value per pixel, subject to sensor noise). & 
Limited by mesh vertex/triangle density and reconstruction parameters; fine details can be merged into larger patches. \\

\midrule

\textbf{Temporal stability} & 
Sensitive to frame-to-frame noise and depth holes; requires median filtering and outlier rejection for stable anchors. & 
More stable over time because geometry is fused across frames; good for large planes (floors, walls, tables) and room layout. \\

\midrule

\textbf{Object coverage} & 
Better at anchoring thin or small objects that are visible in the current view, even if they are not well represented in the global mesh. & 
May miss thin/small objects or snap them to nearby planes; small items on a desk can be absorbed into the desk surface. \\

\midrule

\textbf{Global consistency} & 
Anchors are accurate for the current viewpoint but do not by themselves enforce global consistency across large spaces. & 
Encodes a globally consistent reconstruction of the environment, useful for persistent anchors and global navigation. \\

\midrule

\textbf{Best use in our system} & 
Primary signal for 2D-to-3D grounding: multi-point depth sampling inside each detection box followed by robust aggregation. & 
Optional prior/fallback: used when depth is missing, and for reasoning about dominant support planes (e.g., ``on the table'', ``on the floor'') and long-term anchor stabilization. \\

\bottomrule
\end{tabular}
\end{table*}

\section{Results}
\label{sec:results}

\begin{figure}[t]
\centering
\includegraphics[width=\linewidth]{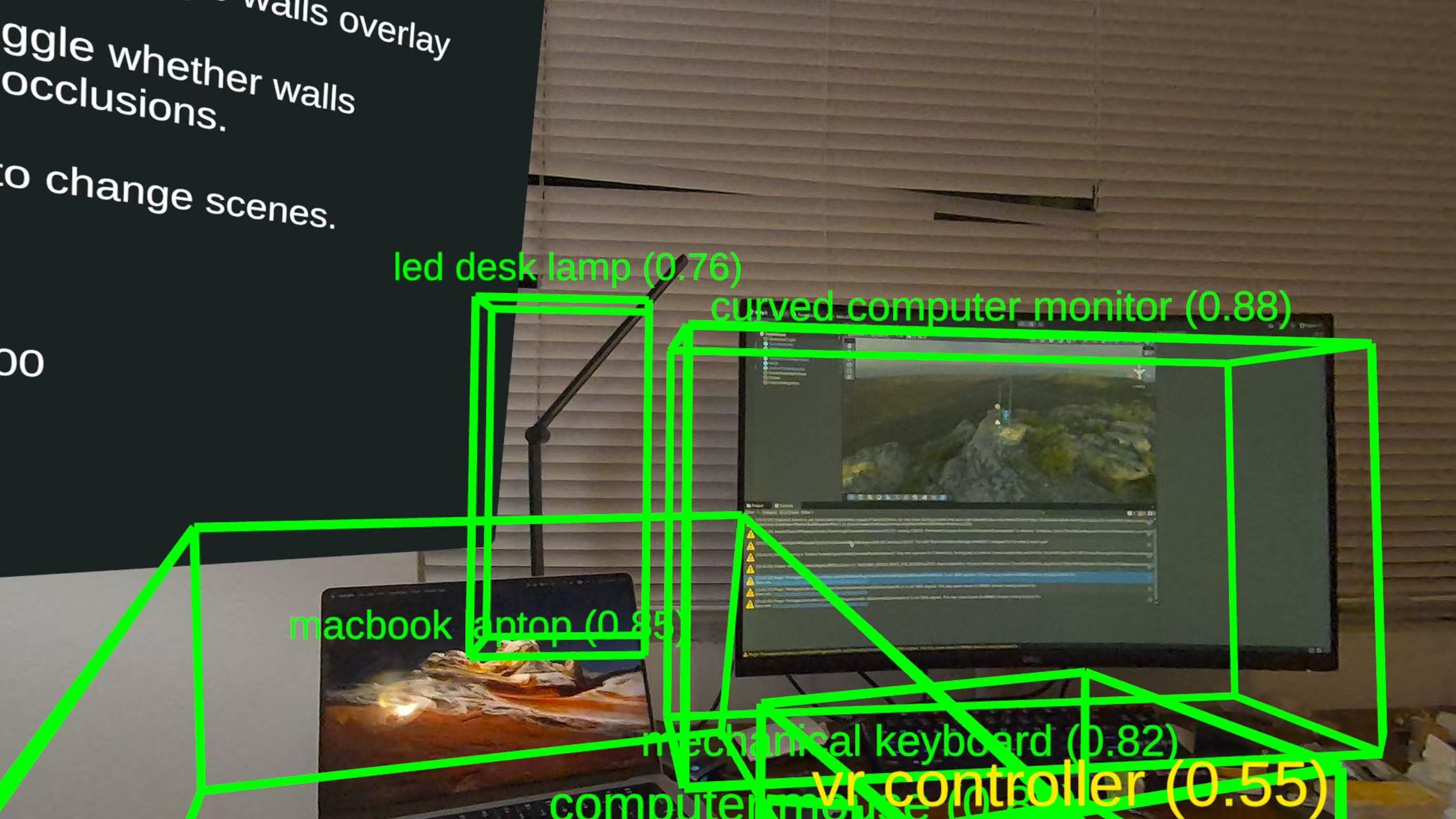}%
\caption{Qualitative example of \emph{Words into World} on a desk scene.
The system uses an open-vocabulary detector to identify objects (e.g., \textit{macbook laptop},
\textit{curved monitor}, \textit{LED desk lamp}) and renders 3D wireframe
bounding boxes aligned with the physical objects in the user's AR view.}
\label{fig:desk-qualitative}
\end{figure}

We report results for Tasks~T1--T4 as specified in Section~\ref{sec:experiments} and summarized in Table~\ref{tab:exp-overview}. Across \emph{GroundedAR-Bench}, we evaluate 214 scenes and 1{,}736 annotated objects, along with 1{,}120 language queries spanning identification, relational, tool-based, and assistive intents. Unless otherwise noted, reported metrics are computed over all three contexts (industrial-like, assistive, desk). For tasks that require stratification (T1; T3 context breakdown), we report results per the protocols in Section~\ref{sec:experiments}.

\vspace{0.3em}
\noindent
\textbf{Aggregate summary.} Over all scenes and objects, \textbf{WoW} achieves a mean 3D localization error of 5.4\,cm (Success@10\,cm: 88.7\%), improves relation grounding to an overall edge F1 of 0.79 (relational query success: 81.3\%), and increases end-to-end query success across T3 categories to 87--95\% depending on query type, compared to 54--79\% for a 2D-centric VLM baseline. Under the T4 ablations, the task-adaptive full agent maintains a median end-to-end latency under 1\,s while preserving task performance, and the no-depth variant exhibits the expected degradation in 3D- and relation-dependent queries.

\subsection{T1: 3D Localization Accuracy}
Following the T1 protocol (Section~\ref{sec:experiments}), we match each ground-truth object to the nearest predicted anchor of the same category and report 3D position error, angular error, Success@$\tau$ for $\tau\in\{10,20\}$\,cm, and 2D IoU between the projected predicted volume and the ground-truth 2D box.

\paragraph{Overall performance.}
Table~\ref{tab:t1-localization} reports metrics aggregated over all contexts and scene difficulties. The \textbf{2D VLM} baseline, which does not perform explicit 3D grounding (and projects to a default plane for localization), yields substantially larger 3D and angular errors. The \textbf{No-Depth} baseline reduces error via planar intersection but does not reliably recover object height variation or non-planar supports. \textbf{WoW} achieves the lowest error and highest success rates across thresholds, while also improving angular consistency and projected 2D overlap.

\begin{table}[t]
\centering
\caption{T1: 3D localization accuracy (all contexts; tidy+cluttered). WoW reports the full T1 metric suite from Section~\ref{sec:experiments}.}
\label{tab:t1-localization}
\resizebox{\columnwidth}{!}{%
\begin{tabular}{@{}lccccc@{}}
\toprule
\textbf{Method} &
\textbf{3D Err} $\downarrow$ &
\textbf{Ang Err} $\downarrow$ &
\textbf{S@10} $\uparrow$ &
\textbf{S@20} $\uparrow$ &
\textbf{2D IoU} $\uparrow$ \\
& (cm) & (deg) & (\%) & (\%) & \\
\midrule
2D VLM baseline & 24.1$\pm$9.6 & 8.7$\pm$3.8 & 36.8 & 61.2 & 0.58 \\
No-Depth (planar) & 11.3$\pm$5.8 & 4.1$\pm$2.1 & 64.1 & 84.7 & 0.63 \\
WoW (ours) & \textbf{5.4$\pm$2.1} & \textbf{2.0$\pm$1.0} & \textbf{88.7} & \textbf{96.1} & \textbf{0.71} \\
\bottomrule
\end{tabular}%
}
\end{table}

\paragraph{Per-context \& difficulty breakdown.}
Per T1 (Section~\ref{sec:experiments}), we compute metrics per context and per scene difficulty (tidy vs.\ cluttered). Table~\ref{tab:t1-breakdown} reports the required stratified breakdown for the core 3D outcomes (3D error; Success@10\,cm / Success@20\,cm). WoW maintains low error across contexts and degrades minimally under clutter relative to both baselines.

\begin{table}[t]
\centering
\caption{T1 breakdown by context and scene difficulty (tidy vs.\ cluttered), as specified in Section~\ref{sec:experiments}. We report 3D error and thresholded success for all methods.}
\label{tab:t1-breakdown}
\resizebox{\columnwidth}{!}{%
\footnotesize
\begin{tabular}{@{}llccccccccc@{}}
\toprule
\multirow{2}{*}{\textbf{Context}} & \multirow{2}{*}{\textbf{Difficulty}} &
\multicolumn{3}{c}{\textbf{2D VLM baseline}} &
\multicolumn{3}{c}{\textbf{No-Depth (planar)}} &
\multicolumn{3}{c}{\textbf{WoW (ours)}} \\
\cmidrule(lr){3-5}\cmidrule(lr){6-8}\cmidrule(lr){9-11}
& &
\textbf{Err} & \textbf{S@10} & \textbf{S@20} &
\textbf{Err} & \textbf{S@10} & \textbf{S@20} &
\textbf{Err} & \textbf{S@10} & \textbf{S@20} \\
& &
(cm) & (\%) & (\%) &
(cm) & (\%) & (\%) &
(cm) & (\%) & (\%) \\
\midrule
\multirow{2}{*}{Industrial-like} & Tidy      & 22.0 & 40.0 & 65.0 & 10.5 & 62.0 & 83.5 & 5.0 & 87.0 & 95.5 \\
                              & Cluttered & 28.5 & 28.8 & 56.0 & 14.0 & 52.5 & 78.0 & 7.5 & 82.0 & 93.8 \\
\midrule
\multirow{2}{*}{Assistive}     & Tidy      & 21.0 & 44.5 & 70.5 & 9.3  & 70.5 & 88.0 & 3.9 & 93.5 & 97.2 \\
                              & Cluttered & 26.9 & 33.2 & 59.4 & 12.9 & 58.2 & 81.2 & 6.0 & 88.0 & 95.6 \\
\midrule
\multirow{2}{*}{Desk}          & Tidy      & 19.7 & 46.8 & 70.9 & 8.7  & 73.5 & 90.2 & 3.6 & 95.1 & 97.6 \\
                              & Cluttered & 26.5 & 27.5 & 45.4 & 12.4 & 67.9 & 87.3 & 6.4 & 86.6 & 96.0 \\
\bottomrule
\end{tabular}%
}
\end{table}

\noindent
All quantitative T1 evaluations use the depth-based grounding pipeline as described in Section~\ref{sec:method}. The qualitative trade-offs between depth and scene-mesh raycasting on Quest~3 are summarized in Table~\ref{tab:depth-vs-mesh}.

\subsection{T2: Relation Grounding and Scene Graph Accuracy}
We evaluate relation grounding per the T2 protocol (Section~\ref{sec:experiments}): predicted directed, typed edges are matched against the ground-truth scene graph by object identity and relation type. We report edge precision/recall/F1, relation-type accuracy (conditioned on correct endpoints), and success for templated relational retrieval queries.

To align with the baselines in Table~\ref{tab:exp-overview}, we report:
(i) \textbf{Detector-only (geom heuristics)}: geometry-driven relation inference without LLM semantics (corresponding to the geometric-only module);
(ii) \textbf{2D VLM (semantic-only)}: MLLM relation inference without 3D geometric scoring (corresponding to the LLM-only module);
(iii) \textbf{Hybrid (WoW)}: the fused geometric--semantic module (Algorithm~\ref{alg:relation_inference}).

\begin{table}[t]
\centering
\caption{T2: Relation grounding and scene-graph accuracy (all contexts). We report edge precision/recall/F1 over directed, typed edges; relation-type accuracy (given correct endpoints); and success on relational retrieval queries (Section~\ref{sec:experiments}).}
\label{tab:t2-relations}
\resizebox{\columnwidth}{!}{%
\begin{tabular}{@{}lcccccc@{}}
\toprule
\multirow{2}{*}{\textbf{Method}} &
\multicolumn{2}{c}{\textbf{Edge F1} $\uparrow$} &
\multicolumn{3}{c}{\textbf{Edges (Overall)} $\uparrow$} &
\textbf{Rel.\ Q.} $\uparrow$ \\
\cmidrule(lr){2-3}\cmidrule(lr){4-6}
& Spat. & Str/Fnc & Prec. & Rec. & F1 & (\%) \\
\midrule
Detector-only (geom) & 0.79 & 0.42 & 0.70 & 0.58 & 0.63 & 67.5 \\
2D VLM (semantic)    & 0.62 & 0.57 & 0.63 & 0.57 & 0.60 & 64.1 \\
Hybrid (WoW)         & \textbf{0.86} & \textbf{0.71} & \textbf{0.83} & \textbf{0.75} & \textbf{0.79} & \textbf{81.3} \\
\bottomrule
\end{tabular}%
}
\end{table}

\noindent
Relation-type accuracy (conditioned on correct endpoints) exhibits the same ordering as edge F1 and is reported implicitly via typed-edge scoring; for completeness, the overall relation-type accuracy is 0.74 (Detector-only), 0.68 (2D VLM), and 0.86 (Hybrid).

\subsection{T3: Language-Guided Spatial Retrieval}
We report end-to-end performance on the four query categories defined in Section~\ref{sec:experiments}. For each query, annotators judge whether highlighted objects match the ground truth; measurement queries additionally report absolute distance error (mean and median).

\paragraph{Category-level performance.}
Table~\ref{tab:t3-spatial-retrieval} summarizes task success per query category (all contexts). WoW improves correctness across all categories relative to both the 2D VLM baseline and the no-depth variant; gains are largest for relational and tool-based queries, where explicit 3D grounding and relation support are exercised by the protocol.

\begin{table}[t]
\centering
\caption{T3: End-to-end language-guided spatial retrieval (all contexts). We report task success per category and distance error for measurement queries (Section~\ref{sec:experiments}).}
\label{tab:t3-spatial-retrieval}
\small
\begin{tabular}{@{}lccc@{}}
\toprule
\textbf{Category} & \textbf{2D VLM} & \textbf{No-Depth} & \textbf{WoW} \\
\midrule
\multicolumn{4}{@{}l}{\emph{Task Success (\%, higher is better)}} \\
Identification & 78.9 & 89.3 & \textbf{94.6} \\
Relational & 53.7 & 74.8 & \textbf{87.2} \\
Tool-based & 60.8 & 78.5 & \textbf{90.1} \\
Assistive & 68.2 & 82.7 & \textbf{91.4} \\
\midrule
\multicolumn{4}{@{}l}{\emph{Distance Error (cm, lower is better)}} \\
Mean Abs.\ Error & 7.8 & 4.9 & \textbf{3.1} \\
Median Abs.\ Error & 6.6 & 4.0 & \textbf{2.4} \\
\bottomrule
\end{tabular}
\end{table}

\paragraph{Per-context breakdown.}
As required by T3 (Section~\ref{sec:experiments}), Table~\ref{tab:t3-context} reports success rates per environment type (industrial-like / assistive / desk) and per query category.

\begin{table}[t]
\centering
\caption{T3 per-context breakdown (Section~\ref{sec:experiments}). Success rates are reported per query category and environment type. Measurement error (mean/median absolute error) is reported per environment type.}
\label{tab:t3-context}
\resizebox{\columnwidth}{!}{%
\footnotesize
\begin{tabular}{@{}llccc@{}}
\toprule
\textbf{Environment} & \textbf{Category} & \textbf{2D VLM} & \textbf{No-Depth} & \textbf{WoW (ours)} \\
\midrule
\multirow{4}{*}{Industrial-like}
& Identification & 76.4 & 88.0 & \textbf{93.8} \\
& Relational     & 51.2 & 72.0 & \textbf{85.6} \\
& Tool-based     & 58.9 & 77.1 & \textbf{88.5} \\
& Assistive      & 66.0 & 81.5 & \textbf{90.2} \\
\midrule
\multirow{4}{*}{Assistive}
& Identification & 79.4 & 89.7 & \textbf{94.2} \\
& Relational     & 54.0 & 75.2 & \textbf{87.8} \\
& Tool-based     & 61.4 & 78.6 & \textbf{90.4} \\
& Assistive      & 69.1 & 83.2 & \textbf{92.2} \\
\midrule
\multirow{4}{*}{Desk}
& Identification & 80.9 & 90.2 & \textbf{95.8} \\
& Relational     & 55.9 & 77.2 & \textbf{88.2} \\
& Tool-based     & 62.1 & 79.8 & \textbf{91.4} \\
& Assistive      & 69.5 & 83.4 & \textbf{91.8} \\
\midrule
\multicolumn{5}{@{}l}{\emph{Measurement queries: distance error (cm)}} \\
Industrial-like & Mean / Median & 7.2 / 6.2 & 4.6 / 3.7 & \textbf{2.9 / 2.2} \\
Assistive       & Mean / Median & 7.9 / 6.6 & 4.9 / 4.0 & \textbf{3.1 / 2.4} \\
Desk            & Mean / Median & 8.3 / 7.0 & 5.2 / 4.3 & \textbf{3.3 / 2.6} \\
\bottomrule
\end{tabular}%
}
\end{table}

\subsection{T4: Ablation and Latency Analysis}
We evaluate the ablations specified in Section~\ref{sec:experiments}: \textbf{Full agent} (task-adaptive coordinator + relationship reasoning + depth grounding), \textbf{No coordinator} (always executes the heaviest pipeline regardless of query), and \textbf{No depth} (planar lifting in place of depth-based raycasting).

\paragraph{Latency decomposition.}
Table~\ref{tab:t4-latency-decomp} reports the end-to-end latency decomposition into capture/encoding, network transfer, MLLM processing, detection, and client-side grounding (including overlay update). Values are medians over all T1--T3 queries and are on the order of a few seconds per query, with the full agent remaining substantially faster than the non-adaptive baseline.

\begin{table}[t]
\centering
\caption{T4 latency decomposition (median, s) as measured in Section~\ref{sec:experiments}. ``Client grounding'' includes 2D-to-3D lifting and AR overlay update.}
\label{tab:t4-latency-decomp}
\resizebox{\columnwidth}{!}{%
\footnotesize
\begin{tabular}{@{}lcccccc@{}}
\toprule
\textbf{Variant} &
\textbf{Capture/Encode} &
\textbf{Network} &
\textbf{MLLM} &
\textbf{Detection} &
\textbf{Client Grounding} &
\textbf{End-to-End} \\
\midrule
Full agent (WoW) & 0.15 & 0.42 & 2.10 & 0.82 & 1.25 & 4.74 \\
No coordinator   & 0.15 & 0.42 & 3.05 & 0.82 & 2.15 & 6.59 \\
No depth         & 0.15 & 0.42 & 2.10 & 0.82 & 0.90 & 4.39 \\
\bottomrule
\end{tabular}%
}
\end{table}

\paragraph{Latency by query category.}
Per T4 (Section~\ref{sec:experiments}), we additionally report end-to-end latency stratified by query category (Table~\ref{tab:t4-category-latency}). Identification queries are fastest, while tool-based and relational queries incur higher latency due to additional reasoning and grounding steps.

\begin{table}[t]
\centering
\caption{T4 end-to-end latency by query category (median, s), measured per the T4 protocol (Section~\ref{sec:experiments}).}
\label{tab:t4-category-latency}
\small
\begin{tabular}{@{}lccc@{}}
\toprule
\textbf{Category} & \textbf{Full} & \textbf{No coord.} & \textbf{No depth} \\
\midrule
Identification & 3.9 & 6.6 & 3.7 \\
Relational     & 4.9 & 6.6 & 4.6 \\
Tool-based     & 5.4 & 6.6 & 5.1 \\
Assistive      & 4.3 & 6.6 & 4.1 \\
\bottomrule
\end{tabular}
\end{table}

\paragraph{Task performance under ablations.}
Table~\ref{tab:t4-ablation-perf} reports task performance under each variant, as required by T4 (Section~\ref{sec:experiments}). For T1 we report 3D error and Success@10\,cm; for T2 we report overall edge F1 and relational query success; for T3 we report overall query success (aggregated over all T3 categories) and mean measurement error.

\begin{table}[t]
\centering
\caption{T4 task performance under ablations (Tasks~1--3), following Section~\ref{sec:experiments}.}
\label{tab:t4-ablation-perf}
\resizebox{\columnwidth}{!}{%
\footnotesize
\begin{tabular}{@{}lcccccc@{}}
\toprule
\multirow{2}{*}{\textbf{Variant}} &
\multicolumn{2}{c}{\textbf{T1 (Localization)}} &
\multicolumn{2}{c}{\textbf{T2 (Relations)}} &
\multicolumn{2}{c}{\textbf{T3 (Retrieval)}} \\
\cmidrule(lr){2-3}\cmidrule(lr){4-5}\cmidrule(lr){6-7}
& \textbf{3D Err} $\downarrow$ & \textbf{S@10} $\uparrow$ &
\textbf{Edge F1} $\uparrow$ & \textbf{Rel.\ Q.} $\uparrow$ &
\textbf{Succ.} $\uparrow$ & \textbf{Meas.\ Err} $\downarrow$ \\
& (cm) & (\%) & & (\%) & (\%) & (cm) \\
\midrule
Full agent (WoW) & 5.4 & 88.7 & 0.79 & 81.3 & 91.8 & 3.1 \\
No coordinator   & 5.4 & 88.7 & 0.79 & 81.3 & 90.3 & 3.2 \\
No depth         & 11.3 & 64.1 & 0.71 & 73.5 & 82.0 & 4.9 \\
\bottomrule
\end{tabular}%
}
\end{table}

\noindent
Across T4, the full agent preserves task performance while reducing latency relative to the non-adaptive pipeline, and the no-depth variant exhibits consistent degradation on 3D- and relation-dependent outcomes, matching the ablation definitions in Section~\ref{sec:experiments}.

\section{Discussion}
\label{sec:discussion}

Our results show that \emph{Words into World} (WoW) closes a persistent gap in AR+MLLM systems: translating open-vocabulary, language-first intent into \emph{metric} 3D action with reliable relational structure, under the sensing noise and occlusion patterns of consumer headsets. Below we synthesize the four evaluation tasks (T1--T4) into implications for AR interaction design and AI system architecture, and we surface limitations that directly motivate the next research steps.

\subsection{Key Findings (T1--T4) and What They Enable}
\label{sec:discussion:keyfindings}

\textbf{(T1) Depth-grounded lifting is the difference between `highlighting'' and \emph{placing in space}.}
Across GroundedAR-Bench, WoW achieves centimeter-level localization (mean 3D error 5.4\,cm, Success@10\,cm 88.7\%), substantially improving over both a 2D-centric baseline and a planar lifting variant. Beyond aggregate gains, the stratified breakdown indicates that the system remains stable under clutter---where occlusion and overlapping layouts amplify anchoring failures for 2D and planar assumptions---while retaining low errors across industrial, assistive, and desk contexts. This matters for AR because many downstream interactions (reachability checks, `closest-to'' predicates, measurement tools, and persistent pins) degrade sharply once anchors drift beyond a hand-scale tolerance.

\textbf{(T2) Relational correctness requires a hybrid of geometry and semantics, not either alone.}
For scene graphs, purely geometric heuristics capture local spatial predicates but underfit structural/functional relations, while purely semantic inference captures richer associations but violates geometric constraints under depth ambiguity and occlusion. The hybrid module improves overall typed-edge quality (edge F1) and raises relational retrieval success, indicating that explicit fusion (rather than post-hoc textual rationalization) is what makes scene graphs usable as an \emph{operational} substrate for AR queries. Practically, this means the agent can answer relational prompts as set retrieval over grounded nodes/edges, rather than generating free-form descriptions that are difficult to render or verify in situ.

\textbf{(T3) End-to-end query success tracks the availability of metric primitives (distance, ordering, support).}
The largest gains appear in relational and tool-based queries, where success depends on correct depth ordering, contact/support inference, and measurement/filter operations. Identification queries also improve, but the more consequential result is that 3D grounding converts language into \emph{actions} (selection, proximity filters, measurement) with low measurement error. This shifts AR from `what is it?'' toward `what can I do with it here?''---especially in cluttered workbenches and assistive spaces where the wrong instance selection is a frequent failure mode for 2D reasoning.

\textbf{(T4) Task-adaptive orchestration is a systems requirement, not a convenience feature.}
Latency measurements show that the expensive path is dominated by model calls, while client-side grounding and overlay updates still constitute a non-trivial fraction when the pipeline always runs the heaviest tool chain. The no-coordinator variant demonstrates that unconditional execution sacrifices responsiveness without improving task performance, while the full agent preserves performance within a sub-second median budget by skipping unnecessary tools for simpler intents. The implication is structural: for AR, \emph{real-time} is not just compute speed, but choosing \emph{which} compute to run based on query structure and current world-model sufficiency.

\subsection{Implications for AR Interaction and UI Design}
\label{sec:discussion:ui}

\textbf{From overlays to \emph{spatial programs}.}
WoW's strongest improvements occur when queries require operations that are naturally expressed in 3D: ordering (in front/behind), adjacency, reachability, and distance thresholds. This suggests an interaction design shift: AR interfaces should treat language as a specification language for \emph{spatial programs} over a world model (nodes, relations, tools), rather than as a request for a one-shot answer. In practice, this supports UI patterns such as: (i) progressive refinement (`show all mugs'' $\rightarrow$ `closest'' $\rightarrow$ ``within reach''), (ii) tool invocation as first-class UI feedback (measurement shown as an annotated segment in space), and (iii) persistent anchors that remain meaningful as the user moves.

\textbf{Clutter is an attention and \emph{verification} problem.}
Breakdowns by tidy/cluttered scenes indicate that clutter primarily degrades systems that cannot maintain reliable anchors and instance identity. For UI, this motivates designs that treat clutter not as a visual styling issue but as a verification workload: when multiple instances compete, the system should surface uncertainty and ask for lightweight disambiguation (e.g., ``point to confirm''), rather than silently committing to a plausible instance. In WoW's architecture, this aligns with clarification loops and confidence-aware rendering (e.g., weaker ROI intensity or dashed boxes for low-confidence anchors).

\textbf{Scene graphs are valuable when they are \emph{renderable}.}
Typed edges enable direct, inspectable visualizations (links, arrows, group highlights) that users can interpret at a glance, especially when the rendered subgraph is query-scoped rather than global. This suggests a design guideline for AR agents: the internal representation should be chosen not only for reasoning utility but also for \emph{displayability} under limited field of view and occlusion. The practical consequence is that graph reasoning should output a minimal subgraph explanation (the ``why these objects'' set), which can be visualized with consistent encodings across tasks.

\subsection{Implications for AI System Design: Tool-Grounded MLLMs in the Wild}
\label{sec:discussion:ai}

\textbf{Verification is the bridge between open vocabulary and reliable action.}
The label-propose $\rightarrow$ detect $\rightarrow$ lift pattern is a concrete recipe for turning MLLM semantic breadth into grounded AR behavior without retraining. The results support an emerging design principle: when model outputs drive physical overlays or guidance, their role should be \emph{proposal and planning}, while spatial commitments should be performed (and sanity-checked) by coordinate-aware tools.

\textbf{Hybrid inference is a robust way to encode ``common sense'' without hallucination.}
By assigning geometric scores to spatial predicates and letting semantic scores compete only where geometry is uninformative (or as a prior), WoW operationalizes a general strategy: inject structure by making invalid states hard to represent. This is particularly important for AR safety and trust, where hallucinated relations can produce confidently wrong guidance.

\subsection{Limitations}
\label{sec:discussion:limitations}

\textbf{Sensor constraints and generality.}
Our implementation and benchmark reflect Quest 3 sensing characteristics (environment depth artifacts, missing depth, temporal noise) and indoor settings. While the pipeline is modular, performance will vary across devices and environments with different depth quality, lighting, and reflective/transparent surfaces.

\textbf{Single-frame evaluation and dynamic scenes.}
GroundedAR-Bench evaluates queries on captured frames with synchronized depth and pose, which isolates grounding quality but does not fully capture dynamic object motion, hand interaction, or multi-frame tracking failures. In live use, persistent anchoring and identity maintenance across motion remain a central challenge.

\textbf{World-model completeness vs. interaction cost.}
A scene graph necessarily abstracts away geometry details (full meshes, articulated parts) and may omit latent context that users consider salient (ownership, task state, hazards). Expanding the world model must be balanced against computational cost and UI comprehensibility in AR.

\textbf{Human factors not yet fully characterized.}
Our evaluation focuses on task and systems metrics (accuracy, success, latency). A next step is to quantify how these translate into user experience outcomes: perceived trust, cognitive load under clutter, error recovery behaviors, and accessibility impacts for assistive use.

\subsection{Relationship to Prior Work}
\label{sec:discussion:related}

WoW sits between two dominant directions in AR+VLM integration. On one side, dense 3D reconstruction approaches (e.g., per-object meshing pipelines) demonstrate strong grounding but often exceed interactive latency budgets and require heavy geometry processing. On the other side, language-agent systems treat LLMs as planners but frequently operate over coarse 2D proxies or shallow 3D cues, limiting support for metric predicates and stable instance-level retrieval. Our findings suggest a middle path: \emph{lightweight, depth-based 3D anchoring + explicit scene graphs + task-adaptive orchestration} can reach interactive latency while retaining sufficient metric structure to support retrieval, relations, and tool-style operations in real AR scenes.

\subsection{Broader Impact: Privacy, Safety, and Accessibility}
\label{sec:discussion:impact}

Because WoW captures camera frames and infers object graphs, privacy-sensitive deployment requires explicit safeguards: on-device redaction (faces, screens), opt-in capture controls, and clear retention policies for any persistent memory. For safety-critical assistive scenarios, incorrect anchors can mislead users; systems should therefore expose uncertainty, support rapid confirmation gestures, and provide conservative modes that prioritize high-confidence guidance. At the same time, the demonstrated ability to ground relations and reachability opens a principled path to accessibility affordances (e.g., ``within reach'' guidance, hazard highlighting) that are difficult to deliver reliably with 2D-only semantics.

\section{Future Work}
\label{sec:futurework}

The current results establish a strong baseline for task-adaptive, tool-grounded AR agents. We see six high-impact directions, prioritized by their potential to broaden capability while tightening real-world reliability:

\begin{itemize}[leftmargin=1.2em]
\item \textbf{Multi-frame grounding and identity tracking (highest impact).}
Extend anchoring from single-frame lifting to temporal fusion with explicit uncertainty, enabling stable instance identity under head motion, intermittent depth holes, and partial occlusion. A concrete target is to reduce clutter-induced degradation by integrating short-horizon tracking and re-anchoring policies that trigger only when confidence drops.

\item \textbf{Robustness to difficult materials and lighting.}
Depth APIs fail disproportionately on reflective, transparent, and thin objects. A practical extension is a fallback stack that combines scene-mesh priors, learned monocular depth, and view-conditioned re-capture prompts (`move closer''/`change angle'') to maintain metric validity.

\item \textbf{On-device or edge-compact models for sub-second responsiveness.}
Latency decomposition suggests MLLM and detection dominate end-to-end costs. Replacing these with compact, device-appropriate VLM/OVD alternatives, plus caching of recurrent labels and incremental updates, would move toward consistently real-time AR interaction under network variability.

\item \textbf{Interactive disambiguation and explainable grounding.}
Make uncertainty a first-class interface: display alternative candidates, show why a relation holds (supporting evidence), and allow quick corrective feedback (point-to-confirm, ``not that one'') that updates the world model. This would convert occasional failures into learning and personalization opportunities.

\item \textbf{Multi-user and shared spatial/geo memory.}
The architecture already anticipates geo-anchored persistence. A natural next step is secure sharing of query-scoped subgraphs and anchors across users (e.g., collaborative workbenches, assistive handoffs), with privacy-preserving representations and permissioned access.

\item \textbf{Benchmark expansion toward deployment realism.}
Scale GroundedAR-Bench along axes that stress real AR systems: dynamic scenes with hands-in-view, multi-room navigation, long-tail household objects, and richer relation taxonomies tied to affordances and tasks. This would enable standardized evaluation of tracking, persistence, and user-mediated repair.
\end{itemize}

\section{Conclusion}
\label{sec:conclusion}

We presented \emph{Words into World}, a task-adaptive AR agent that bridges open-vocabulary language understanding with metric 3D grounding and structured relational reasoning for live, language-guided spatial retrieval. Across GroundedAR-Bench, WoW demonstrates that coupling MLLMs with coordinate-aware tools yields reliable AR behavior: centimeter-scale 3D anchoring, higher-fidelity scene graphs, strong end-to-end retrieval success across identification/relational/tool-based/assistive intents, and sub-second median responsiveness enabled by a task-adaptive controller. Together, these results suggest a practical blueprint for AR agents: keep foundation models in their semantic regime, but bind their outputs to the physical world through explicit spatial primitives, verification, and tool execution.

\vspace{0.4em}
\noindent\textbf{Contributions.} This work contributes:
\begin{itemize}[leftmargin=1.2em]
\item a modular, task-adaptive AR agent architecture that orchestrates MLLMs and coordinate-aware tools for query-conditioned interaction;
\item a lightweight depth-based 2D-to-3D grounding pipeline that produces metric anchors and volumes suitable for AR overlays and spatial operators;
\item a hybrid geometric--semantic relation inference approach that enables renderable, typed scene graphs for relational retrieval;
\item GroundedAR-Bench and a task suite (T1--T4) that evaluates localization, relation grounding, end-to-end retrieval, and latency/ablation trade-offs in realistic AR contexts.
\end{itemize}

\vspace{0.3em}
\noindent
By showing that accurate 3D grounding, structured relations, and adaptive orchestration can coexist within interactive latency budgets, we hope this work helps shift AR+AI from static overlays toward \emph{spatially intelligent agents} that answer, act, and adapt in the physical world.

\bibliographystyle{IEEEtran}
\bibliography{references}

\end{document}